\documentclass[10pt,twocolumn,letterpaper]{article}

\usepackage{cvpr}
\usepackage{times}
\usepackage{epsfig}
\usepackage{graphicx}
\usepackage{amsmath}
\usepackage{amssymb}
\usepackage{placeins}
\usepackage{tabularx}

\newcommand{\fig}[1]{Figure~\ref{fig:#1}}

\newcommand{\eq}[1]{(\ref{eq:#1})}

\newcommand{\e}{e}

\usepackage[pagebackref=true,breaklinks=true,letterpaper=true,colorlinks,bookmarks=false]{hyperref}

\cvprfinalcopy % *** Uncomment this line for the final submission

\def\cvprPaperID{9579} % *** Enter the ICCV Paper ID here

\ifcvprfinal\fi
\begin{document}

%%%%%%%%% TITLE
\title{Instance Segmentation of Biological Images Using Harmonic Embeddings}

\author{Victor Kulikov\\
PicsArt Inc.\thanks{Work done while at Skolkovo Institute of Science and Technology.}\\
Moscow, Russian Federation\\
{\tt\small victor.kulikov@picsart.com}
\and
Victor Lempitsky\\
Samsung AI Center Moscow\\
Skolkovo Institute of Science and Technology\\
{\tt\small lempitsky@skoltech.ru}
}

\maketitle
\begin{abstract}
We present a new instance segmentation approach tailored to biological images, where instances may correspond to individual cells, organisms or plant parts. Unlike instance segmentation for user photographs or road scenes, in biological data object instances may be particularly densely packed, the appearance variation may be particularly low, the processing power may be restricted, while, on the other hand, the variability of sizes of individual instances may be limited. The proposed approach successfully addresses these peculiarities.

Our approach describes each object instance using an expectation of a limited number of sine waves with frequencies and phases adjusted to particular object sizes and densities. At train time, a fully-convolutional network is learned to predict the object embeddings at each pixel using a simple pixelwise regression loss, while at test time the instances are recovered using clustering in the embedding space. In the experiments, we show that our approach outperforms previous embedding-based instance segmentation approaches on a number of biological datasets, achieving state-of-the-art on a popular CVPPP benchmark. This excellent performance is combined with computational efficiency that is needed for deployment to domain specialists.

The source code of the approach is available at \texttt{https://github.com/kulikovv/harmonic}~.

% At test time, a deep convolutional network predicts special kind of embeddings for each pixel, so that pixels of the same instances have similar embeddings and object instances can be recovered using clustering in the embedding space. Unlike previous works, the 
% The biological object in microscopy are often heavily occluded, and bounding box based methods like Mask-RCNN may fail to parse overlapping objects of the same class. In this work, we tried to solve the problem of parsing overlapping biological objects with a new instance embedding framework: harmonic instance segmentation.
% The main idea of the proposed framework is the implicit embedding of each pixel as the expectation over a range of harmonic functions. Our approach allows us to define the embedding space at training time and to train a feedforward convolutional network to minimize the distance between ground truth embedding and neural network output in an end-to-end fashion.  To make this learning process possible, we have introduced a SinConv layer, that propagates the underline harmonic function space in the decoding layers of the network. At the test time, a clustering algorithm is applied to retrieve instances from the embedding space. 
% In the experimental validation, the harmonic instance embedding approach is shown to be capable of solving different instance parsing tasks. We have demonstrated that our method reaches the state-of-the-art on three biological datasets: plant phenotyping (CVPPP), roundworms and HeLa cells datasets.   

%The source code will be publicly available.  
\end{abstract}
\section{Introduction}
% \begin{figure*}[t]
% \begin{center}
% \includegraphics[width=\textwidth,keepaspectratio]{figures/Teaser.pdf}
% \end{center}
% \caption{ \textbf{Harmonic instance segmentation} framework. At test time the input image $I$ is processed by encoder/decoder fully convolutional neural network $\Phi_\theta$. At each up-sampling harmonic embedding function $F$ are concatenated with features map. The neural network produces embeddings $\Phi_\theta(I,F)$ that are parsed into instance using clustering algorithm. At training time (blue section) the ground truth instance segmentation $G_I$ is converted into ground truth embeddings $E_F(G_I)$ using harmonic embedding function $F$. The network is trained minimizing $|E_F(G_I)-\Phi_\theta(I,F)|$. }
% \label{fig:teaser}
% \end{figure*}

Instance segmentation (object separation) in biological images is often a key step in their analysis. Many biological image modalities (e.g.,\ microscopy images of cell cultures) are characterized by excessive numbers of instances. Other modalities (e.g.,\ worm assays) are characterized by tight and complex overlaps and occlusions. On the other hand, in most situations the scale variations of objects of interest in biomedical data are less drastic than in natural photographs due to the lack of strong perspective effects. In this work, we propose a new approach designed for biological image instance segmentation that can address the challenges (number of instances, overlaps) while exploiting the simplifying properties (limited scale variation). 

\begin{figure*}[t]
\begin{center}
\includegraphics[width=\textwidth,keepaspectratio]{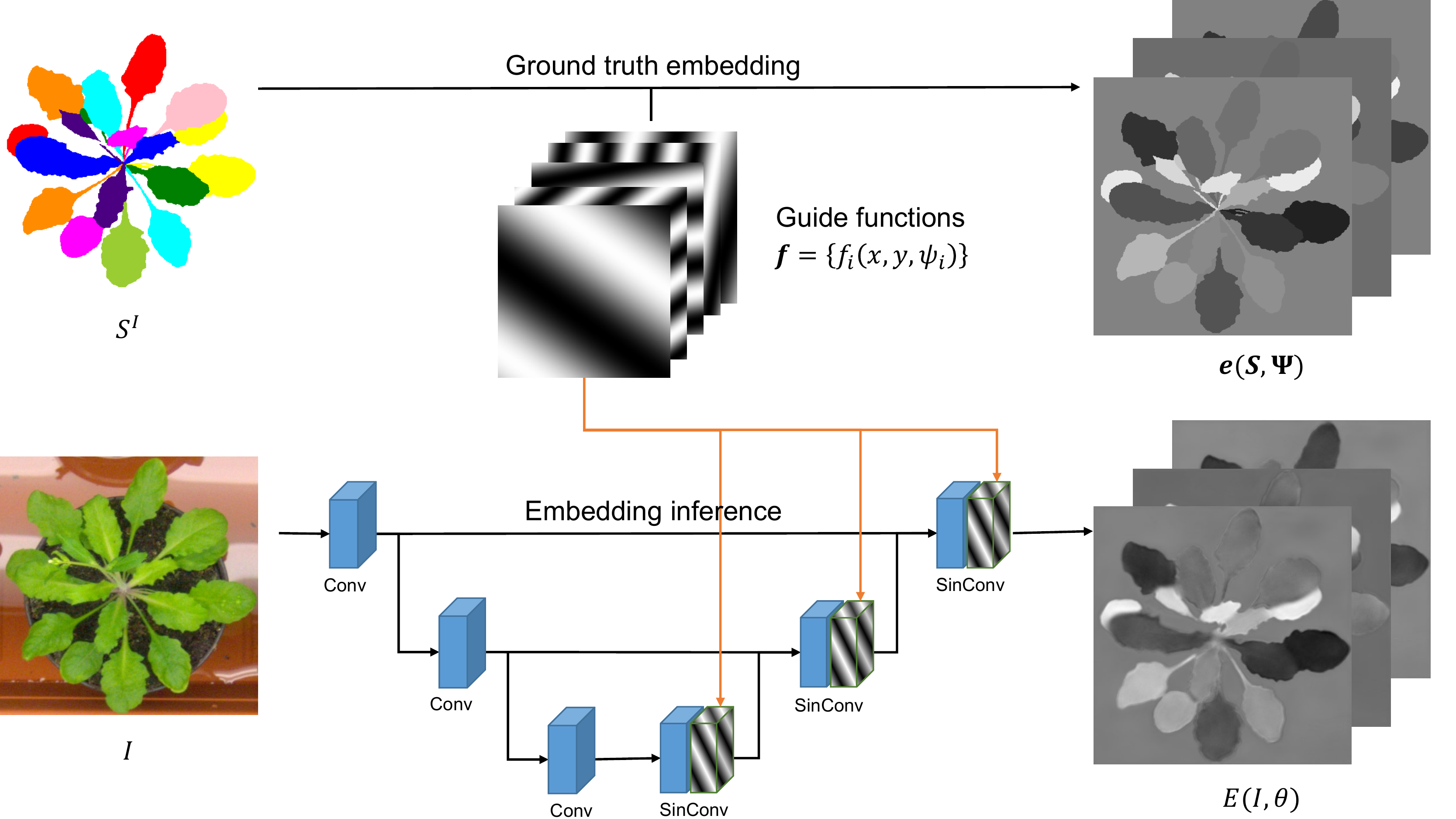}
\end{center}
\caption{The \textbf{harmonic instance segmentation} framework. At train-time, we embed each pixel of the ground truth image $S^I$ as the mean of predefined guide functions $\textbf{f}$ over instance pixels it belongs to, resuling in embeddings $e(S,\Psi)$.  We then train the neural network $E$ to reproduce the ground truth embedding given the inputted image $I$. To simplify learning, guide functions $\textbf{f}$ are inputted into intermediate representations of the network using \textit{SinConv} layers. The learning process uses a simple pixelwise L1-Loss between ground truth embedding $e(S,\Psi)$ and the neural network prediction $E(I,\theta)$ as a learning objective. At test time instances are retrieved from the predicted embedding $E(I,\theta)$ using mean shift clustering.  }
\label{fig:Pipeline}
\end{figure*}

Our approach continues the series of studies~\cite{de2017semantic,novotny2018semi} that perform instance segmentation by learning deep embeddings and using clustering in the embedding space to recover the instances at test time. While this approach is appealing in its simplicity, learning good embeddings for object instances with a fully-convolutional network is  challenging, especially for biological data, where individual instances may have an almost indistinguishable appearance.

In order to utilize the specific nature of biomedical images, we depart from the end-to-end philosophy of the previous embedding-based instance segmentation works and split the learning process into two stages. At the first stage, we seek a small set of harmonic functions that can be used to separate objects in the training dataset. The search is implemented as an optimization process that tunes the frequencies and phases of the harmonics to a specific range of scales and object densities in the data. The selected set of harmonics then guides the second stage of the learning process as well as the inference process at test-time.

At the second learning stage, we assign each ground truth object instance its \textit{harmonic embedding} based on the expectation of the learned set of functions. We then learn a deep fully-convolutional network to predict resulting embeddings at each pixel. We show that learning with a simple pixelwise regression loss is feasible as long as the information about harmonic functions is provided to convolutional layers in the network (which we achieve by augmenting the input to these convolutional layers). The learned networks generalize well to new images, and tend to predict pixel embeddings that can be easily clustered into object instances.

In the experiments, we compare our approach to direct embedding-based instance segmentation~
\cite{de2017semantic} as well as to other state-of-the-art methods. Four biomedical datasets corresponding to plant phenotyping, bacterial and human cell culture microscopy, as well as C.elegans assays, are considered. We observe considerable improvement in performance brought by our approach.

\section{Related Work}

\textbf{Proposal-based instance segmentation} methods~\cite{cordts2016cityscapes,hariharan2014simultaneous,chen2015multi,cordts2016cityscapes,dai2016instance,li2016fully,he2017mask} combine object detection with object mask estimation and currently represent state-of-the-art in instance segmentation non-biological benchmarks. The necessity to perform object detection followed by non-maximum suppression makes learning and tuning of methods from this group complex, especially in the presence of tight object overlaps when non-maximum suppression becomes fragile.

%build pixel-wise masks for each object detection. The most recent approach is to made a network consisting of two parts: bounding box regression head and a semantic segmentation head over ROI-pooling operation. The non-maxima suppression post-processing step is required for the selection of the most reliable detections. Those methods achieve state-of-the-art performance on most challenging datasets.

Another group of approaches to instance segmentation is based on \textbf{recurrent neural networks}~(RNNs) that generate instances sequentially. Thus, Romera~et~al.~\cite{romera2016recurrent} trains a network for end-to-end instance segmentation and counting using LSTM networks~\cite{hochreiter1997long}.  Ren~et~al.~\cite{ren2016end} proposed a combination of a recurrent network with bounding box proposals. RNN-based frameworks show excellent performance on small datasets; they achieve state-of-the-art results on the CVPPP plant phenotyping dataset. The major problem of recurrent methods is the vanishing gradient effect that becomes particularly acute when the number of instances is large.   

%The \textbf{proposal-free methods} involve a kind of metric learning over semantic segmentation followed by a post-processing step.  Usually, those methods predict auxiliary information in order to assist the post-processing. Thus, deep watershed transform~\cite{bai2016deep} learns to predict for each pixel of the object direction and distance to the nearest edge. In \cite{uhrig2016pixel} a template matching algorithm uses semantic segmentation, depth, and angle estimations for each pixel for instance prediction.  

Our method falls into the category of \textit{proposal-free} approaches to instance segmentation based on \textbf{instance embedding}. In this case, neural networks are used to embed pixels of an image into a hidden multidimensional space, where embeddings for pixels belonging to the same instance should be close, while embeddings for pixels of different objects should be separated. A clustering algorithm may then be applied to separate instances. To achieve this, the previous approach~\cite{fathi2017semantic} penalizes pairs of pixels using a logistic distance function in the embedding space. The embedding is learned using log-loss function and requires to weight pixel pairs in order to mitigate the size imbalance issue. This method also predicts a seedness score for each pixel, which correlates with the centeredness. They use this score to pick objects from the embedding. Kong~at~al.~\cite{kong2018recurrent} use differentiable Gaussian Blurring Mean-Shift for the recurrent grouping of embeddings. Deep Coloring~\cite{kulikov2018instance} proposes a reduction of instance segmentation to semantic segmentation, whereas class labels are reused for non-adjacent objects. The instances are then retrieved using connected component analysis. 

In their study most related to ours, De~Brabandere~et~al.~\cite{de2017semantic} use a non-pairwise discriminative loss function composed of two parts: one  pushing the embedding mean of different objects further apart, while the other one pulling the embeddings of the same object pixels closer to its mean. Instances are retrieved using the mean-shift algorithm. The approach~\cite{novotny2018semi} uses metric learning together with an explicit assignment of  the center of mass as the target embedding. Our approach follows the general paradigm of  \cite{de2017semantic, novotny2018semi} but suggests a special kind of embeddings detailed below. The use of new embeddings results in explicit assignment of embeddings to each pixel in the training image, thus simplifying the learning process.

%but instead of learning means of embeddings, we can estimate them using harmonic functions priors. That simplifies the training process, allowing to minimize the distance between the ground-truth embeddings and the embeddings produced by the network and also reduces the dimensionality of the embedded space. We also use mean-shift algorithm to retrieve instances like in \cite{de2017semantic} and instance weighting function are similar to \cite{fathi2017semantic, kulikov2018instance}. 
\section{Harmonic Instance Embedding}
\def\f{\mathbf{f}}
\def\e{\mathbf{e}}
\def\P{\mathcal{P}}
\def\I{\mathcal{I}}
\def\S{\mathbf{S}}

We now discuss our approach in detail.  Existing instance embedding methods \cite{de2017semantic,kulikov2018instance,fathi2017semantic} do not prespecify target embeddings for pixels in the training set. Instead, they rely on the learning process itself to define these embeddings. In contrast, our goal is to define ``good'' embeddings to pixels a priori. ``Goodness'' here means amenability for clustering as well as learnability by a convolutional architecture.

Let $\f=\{f_1(x,y;\psi_1),f_2(x,y;\psi_2),\dots f_N(x,y;\psi_N)\}$ be a family of $N$ real-valued functions $f_i(x,y;\psi_i)$ in the image domain, where $(x,y)$ corresponds to the coordinates of the argument, and $\psi_i$ is a set of learnable parameters defining the shape of the function (e.g.,\ the frequency vector and the phase of a sine function). As our approach is built in many ways around this family of functions, we call the function family $\f$ the \textit{guide functions}.

Let $\S$ be an arbitrary set of pixels (e.g.,\ an object instance in the ground truth annotation of a training image). We denote  the expectation of $f_i$ over $S$ 
with $e_i(\S;\psi_i)$:
\begin{equation} \label{eq:embed}
    e_i(\S;\psi_i) = \frac{1}{|\S|}\sum_{(x,y)\in \S} f_i(x,y;\psi_i)\,. 
\end{equation}
If $\Psi=\{\psi_1,\dots\psi_N\}$ denotes the joint vector of parameters of all $N$ functions, then the \textit{guided embedding} of an object $\S$ determined by $\Psi$ is defined as the following $N$-dimensional vector:
\begin{equation} \label{eq:embedvec}
\e(\S;\Psi) = \{e_1(\S;\psi_1),e_2(\S;\psi_2),\dots,e_N(\S;\psi_N)\}\,.
\end{equation}
To sum up, the guided embedding maps each object $S$ to the vector of the guide functions' expectations over this object.

\subsection{Picking good guide functions}

Given a new dataset representing a new type of instance segmentation problem, our goal is to find a good set of guide functions \eq{embed} so that different objects have well-separated guided embeddings. 

To do that, we first restrict $f_i$ to a certain functional family parameterized by the parameters $\psi_i$. As discussed above, in many biomedical datasets, there is a certain (imperfect) regularity in the location of objects. For example, monolayer cell cultures organize themselves in a texture composed of elements having approximately the same size and being adjacent to each other. Such loosely-regular, semi-periodic structure calls for the use of harmonic functions as guides:
\begin{equation}
f_i(x,y;\psi_i) = \sin\left(\frac{\psi_i[1]}{W}\,x+\frac{\psi_i[2]}{H}\,y+\psi_i[3]\right)\,,
\end{equation} where $W$ and $H$ are image width and height, respectively; $\psi_i[1]$ and $\psi_i[2]$ are frequency parameters; and $\psi_i[3]$ is the phase parameter.

Assume now that a set of training images $\I$ is given. We can then estimate the quality of guided embeddings by looking at pairs $\S$ and $\S'$ of objects belonging to the same image (e.g.,\ two different cells from the same image) and finding out how frequently they have very close embeddings. Ideally, we want to avoid such collisions in the embedding space (at the very least, we want to avoid them in the training set). The following loss is therefore considered:
\begin{align} \label{eq:objemb}
    \ell(\Psi)=\sum_{I\in \I}\frac{1}{|\P_I|}&\sum_{(S,S')\in \P_I}\\\max
&\left(0,\,\epsilon-\left\|\e(\S;\Psi)-\e(\S';\Psi)\right\|_1\right)\,,\nonumber
\end{align}
where $\|{\cdot}\|_1$ is the $L_1$ distance; $\epsilon$ is the margin meta-parameter; and $\P_I$ denotes the set of all pairs of objects from the training image $I$. Each individual term in \eq{objemb} is a hinge loss term that is non-zero if the guided embeddings of a certain object pair are too close (closer than $\epsilon$).

To find good guide functions, we minimize the loss \eq{objemb} on the training set. We perform stochastic gradient descent over the training set by drawing minibatches of random pairs of objects from random images and updating $\Psi$ to minimize \eq{objemb} for the pairs from the minibatch. In our implementation, we initialize frequency parameters $\psi_i[1]$ and $\psi_i[2]$ to uniformly distributed random numbers from the interval $(0;50)$, while the phase parameters $\psi_i[3]$ are initialized uniformly from $[0;2\pi)$.

The outcome of the learning is a set of guide functions, such that pairs of objects from training images have their guided embeddings separated in the embedding space. For typical settings $N=12$ and $\epsilon=0.5$, most pairs in the training set end up being isolated by more than the margin.

\begin{figure}
    \centering
    \includegraphics[width=0.45\textwidth,keepaspectratio]{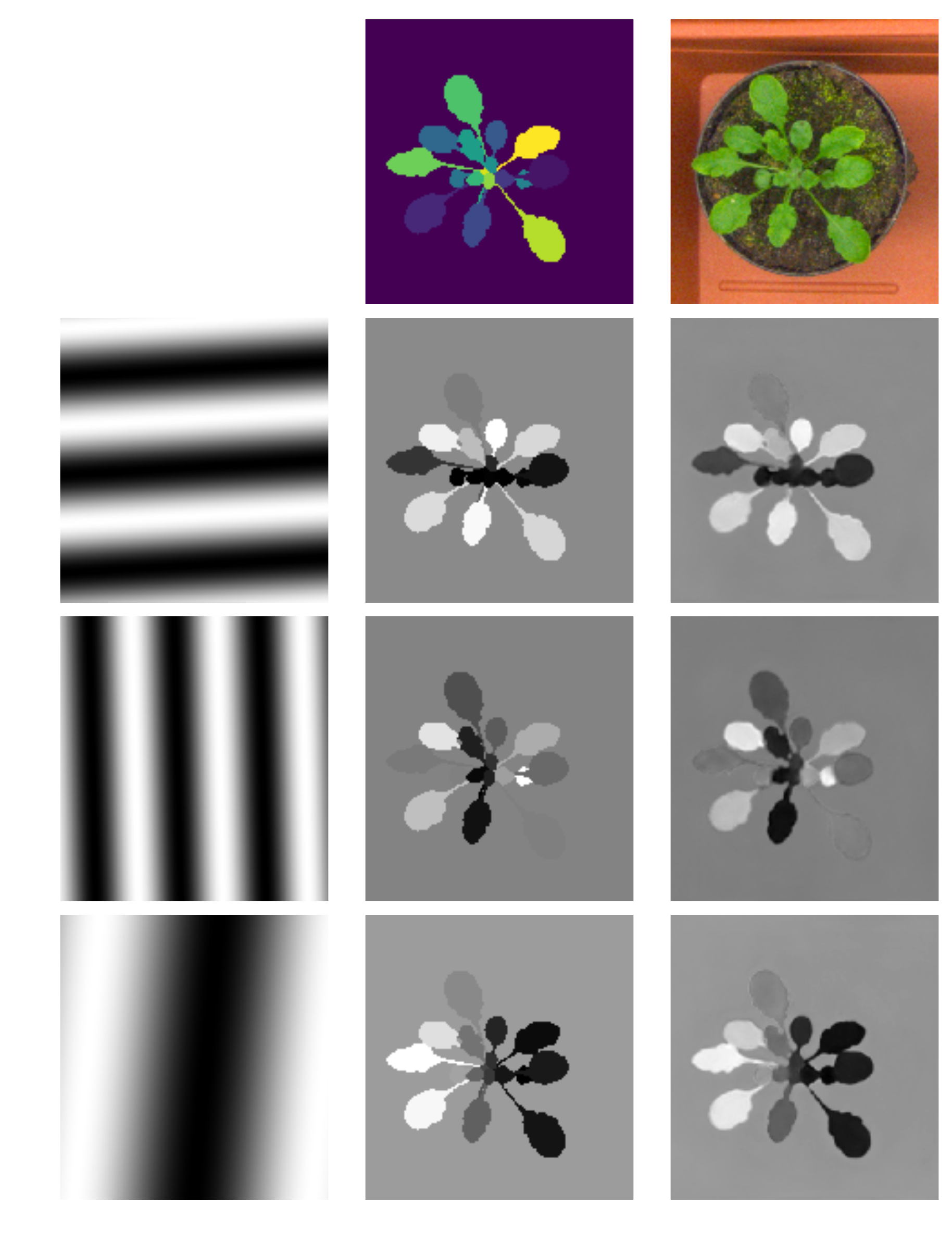}
    \caption{Left -- guide functions (three of them shown here; 12 are used in our experiments). Middle -- the GT masks (top) and the resulting guided embeddings. Note that each instance (leaf) is separated from other leaves in at least some of the guides. Right -- the image (top) and the pixel embeddings as predicted from it by the convolutional network. Embeddings predicted by the network are very close to the guided embeddings, which greatly simplifies the  clustering-based post-processing.
    }
    \label{fig:embeddins}
\end{figure}

\subsection{Learning good embedding network}

Assume now that the parameters of the guide functions have been optimized on the training set so that the parameters $\Psi$ are now fixed. To derive the loss for the second stage, we denote % we further denote $S^I(x,y)$ be a mapping from pixel $(x,y)$ to an object containing this pixel.
$S^I(x,y)$ the set of all pixels of an object containing pixel $(x,y)$.

We then train a deep fully-convolutional embedding network $E(I;\theta)$ with parameters $\theta$ to map input images to sets of $N$-channel images, where each pixel is assigned an $N$-dimensional embedding. We minimize a simple pixelwise loss function that encourages the network to map each pixel to the guided embedding of the corresponding object: 
\begin{equation} \label{eq:netloss}
    \ell(\theta)=\sum_{I\in\I} \sum_{(x,y)\in\text{fg}(I)}\left\|E(I;\theta)[x,y] - \e(S^I(x,y);\Psi)\right\|_1\,.
\end{equation}
Here, $\text{fg}(I)$ denotes the set of foreground pixels of image $I$ and $E(I;\theta)[x,y]$ denotes the output of the network $E$ at the spatial position $(x,y)$ (if the foreground/background segmentation is not available, then the summation is taken over the full image). 

We have found that standard fully-convolutional architectures (e.g.,\ U-Net~\cite{ronneberger2015u}) perform very well (\fig{embeddins}). Such architectures achieve low train and test set losses \eq{netloss} provided one important modification to convolutional layers is made. When modifying a convolutional layer $L$, we augment its input with an extra set of maps holding the guide functions values. Specifically, the extra maps contain the values $\{f_1(x{\cdot}\Delta_L,y{\cdot}\Delta_L;\psi_1),\dots f_N(x{\cdot}\Delta_L,y{\cdot}\Delta_L;\psi_N)\}$ at each spatial position $(x,y)$. Here, $\Delta_L$ is the downsampling factor of the layer $L$ (compared to the input/output resolution). The use of downsampling factor is needed to make sure that the augmenting maps in different layers are spatially aligned with the output. 

\begin{figure}[h]
\begin{center}
\includegraphics[width=0.45\textwidth,keepaspectratio]{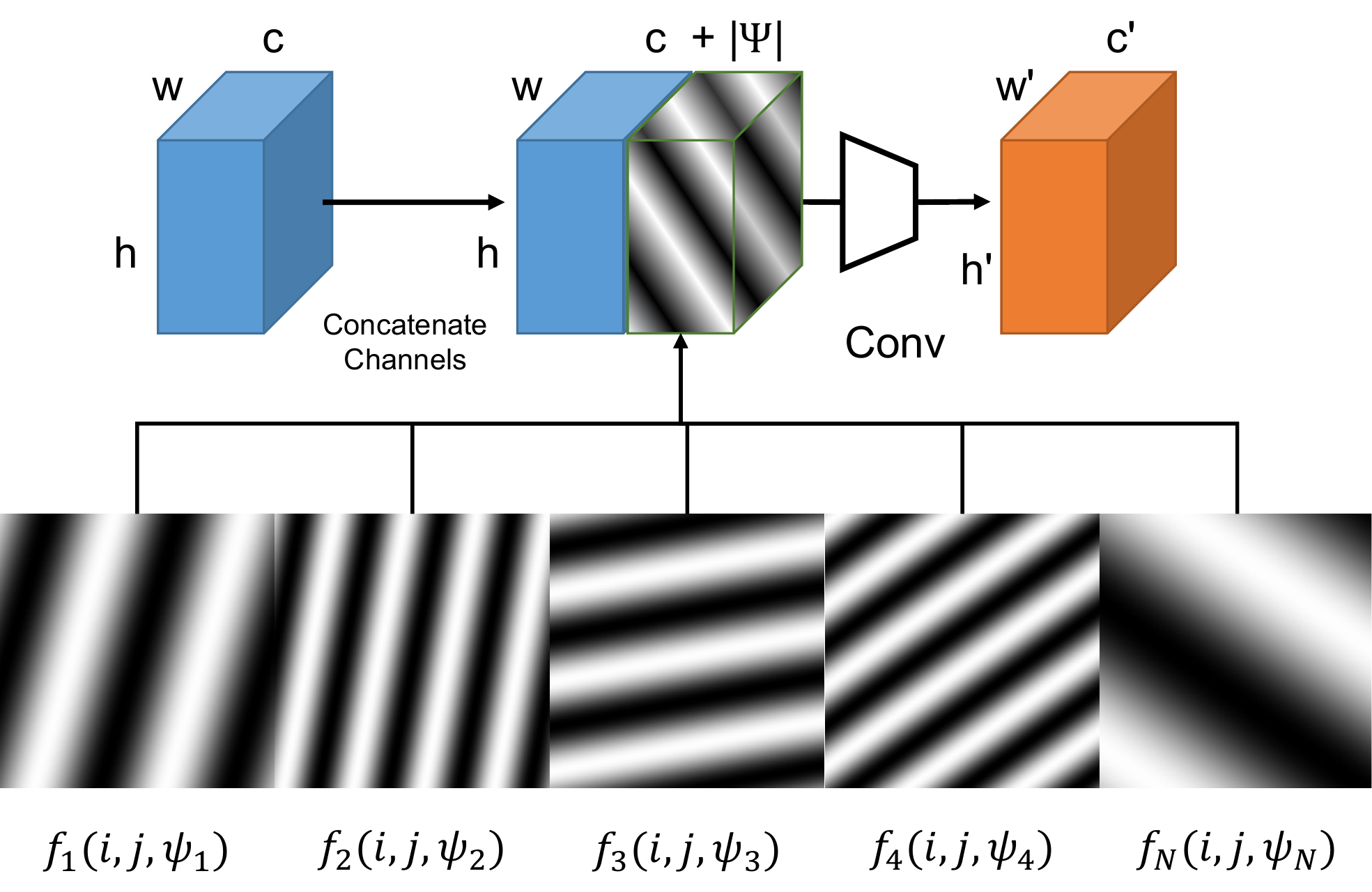}
\end{center}
\caption{ The SinConv layer maps a representation block with shape $w \times h \times c$ to a new representation block of shape $w' \times h' \times c'$ by concatenating guide functions and then performs convolution. The use of SinConv blocks greatly simplifies the task of learning to regress harmonic embeddings.}
\label{fig:SinConv}
\end{figure}

Note that our augmentation idea generalizes the recently suggested CoordConv layer \cite{liu2018intriguing} that augmented the input of convolutional layers with $\{f_1(x,y)=x, f_2(x,y)=y\}$. By analogy and since the guiding functions in our implementation are harmonic, we call the new operation the \textit{SinConv} layer (Figure \ref{fig:SinConv}).

As mentioned above, our embedding architectures follow the design principles of the state-of-the-art semantic segmentation frameworks \cite{ronneberger2015u,zhao2016pyramid,he2016deep} which are composed of encoder and decoder parts with skip-connections between them. We use SinConv layers in the upsampling part (``decoder'') only. This allows to use a pretrained ``backbone'' networks in the encoder part, though we have not used this option in most of our experiments.

\subsection{Instance segmentation of test images}

At test time, the application of the learned embedding network is straightforward. The network $E(I;\theta)$ is applied to an input image.
Our post-processing is then similar to the one suggested in \cite{de2017semantic}. We use the mean-shift clustering algorithm \cite{comaniciu2002mean} to obtain instance masks from the embedding space (Figure \ref{fig:embeddins}, right column). The mean-shift bandwidth is set to the margin $\epsilon$ used in the guide function selection objective \eq{objemb}, since both parameters have the meaning of the desirable separation between the embeddings of different instances.

\section{Experiments}
We provide results of our method on three challenging biomedical datasets of bright-field microscopy images of C.elegans, E.coli, Hela and the plant phenotyping dataset (CVPPP 2017 sequence A1). In each case, learning was done on a single NVidia Tesla V100 GPU using ADAM optimizer with learning rate 1e-5. Our implementation uses PyTorch~\cite{pytorch}. 

The architecture and data augmentation were the same for all datasets. In our experiments, we have used the U-Net \cite{ronneberger2015u} neural network and replaced the first convolution of each upscaling block with the SinConv layer. The network was trained from scratch. Due to a small number of training images in those datasets, we have added some data augmentation procedures, namely, cropping patches of size $448\times448$, scaling, and left-right flips. The number of embedding dimensions was set to 12 and the margin $\epsilon$ was set to $0.5$. With these settings the objective \ref{eq:objemb} is minimized to zero during guide function search. Note that the availability of parameters that work well for diverse datasets is very important for practitioners. The number of training epochs was set differently for different datasets due to their varying complexity. 

We used Symmetric best Dice coefficient (SBD) and average precision (AP) as metrics. The SBD metric averages the intersection over union (IOU) between pairs of predicted and the ground truth labels yielding maximum IOU. The AP metrics integrates precision for different recall values.

We have used the study by   De~Brabandere~et~al.~\cite{de2017semantic} as the main baseline and have reimplemented their approach using the same network architecture as ours (the variants with and without SinConv layers were tried). On the CVPPP dataset where the authors' implementation results are known, the result of our reimplementation is considerably better suggesting that it forms a strong baseline.

\subsection{CVPPP dataset}
\begin{figure*}
\begin{center}
\includegraphics[width=\textwidth,keepaspectratio]{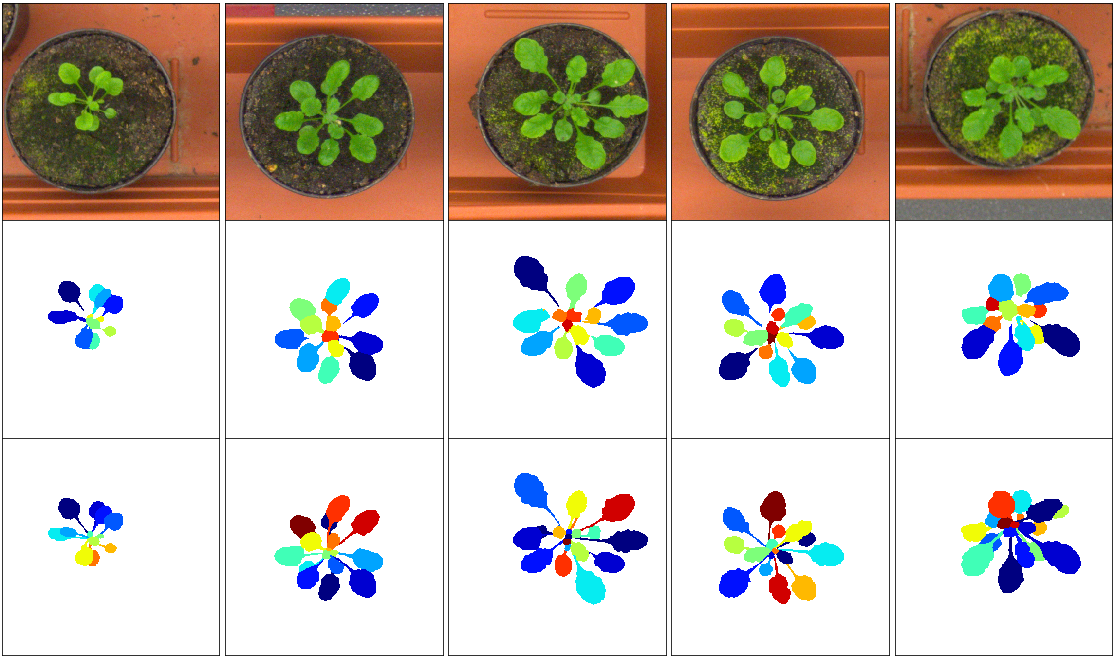}
\end{center}
\caption{ Sample results for the plant phenotyping dataset (CVPPP). The top row shows the source images; the second row shows our results; and the third row shows the ground truth (color display required).}
\label{fig:CVPPP}
\end{figure*}

The Computer Vision Problems in Plant Phenotyping (CVPPP) dataset \cite{scharr2014annotated}(Figure \ref{fig:CVPPP}) is one of the most popular instance segmentation benchmarks. The dataset consists of five sequences of different plants. We have used the most common sequence A1 that has the most significant number of baselines. The A1 sequence has 128 top-down view images, with a size $530\times500$ pixels each, as a training set and an additional hidden test set with 33 images from the same sequence. The task of instance segmentation is challenging because of the high variety of leaf shapes and complex occlusion between leaves. The performance of competing algorithms is SBD metrics and the absolute difference in counting $|DiC|$ (c.f.~\cite{scharr2016leaf}). 

We have trained the neural network for 500 epochs to fit this embedding space. Table \ref{tab:cvppp} shows our method currently being the state-of-the-art in the main SBD metric compared to all  published methods at the moment of submission.  

\begin{table}
    \begin{center}
        \begin{tabularx}{0.45\textwidth}{|X|c c|}
        \hline
        \textit{Method} &$\mid DiC\mid$&$SBD(\%)$\\
        \hline
        IPK \cite{pape20143}&2.6&74.4\\
        Nottingham \cite{scharr2016leaf}&3.8&68.3\\
        MSU \cite{scharr2016leaf}&2.3&66.7\\
        Wageningen \cite{yin2014multi}&2.2&71.1\\
        PRIAn \cite{giuffrida2016learning}&1.3&-\\
        Recurrent IS \cite{romera2016recurrent}&1.1&56.8\\
        Recurrent IS+CRF \cite{romera2016recurrent}&1.1&66.6\\
        Recurrent with attention \cite{ren2016end}&0.8&84.9\\
        Discriminative loss \cite{de2017semantic} &1.0&84.2\\
        Deep coloring \cite{kulikov2018instance}&2.0&80.4\\
        \hline
        \multicolumn{3}{|l|}{Discriminative loss \cite{de2017semantic} (our implementation)}\\
        \hline
        Without SinConv &4.&88.0\\
        With SinConv&4.&89.0\\
        \hline
        Ours without SinConv&5.&78.3\\
        \textbf{Ours}&3.&\textbf{89.9}\\
        \hline
        \end{tabularx}
    \end{center}
     \caption{Quantitative results on the CVPPP dataset (methods with published descriptions as well as our method and baselines included). Our method performs best in terms of  using the Symmetric Best Dice coefficient ($SBD$) compared to previously published approaches.}
    \label{tab:cvppp}
\end{table}

\subsection{E. coli dataset}

\begin{figure}
\begin{center}
\includegraphics[width=0.47\textwidth,keepaspectratio]{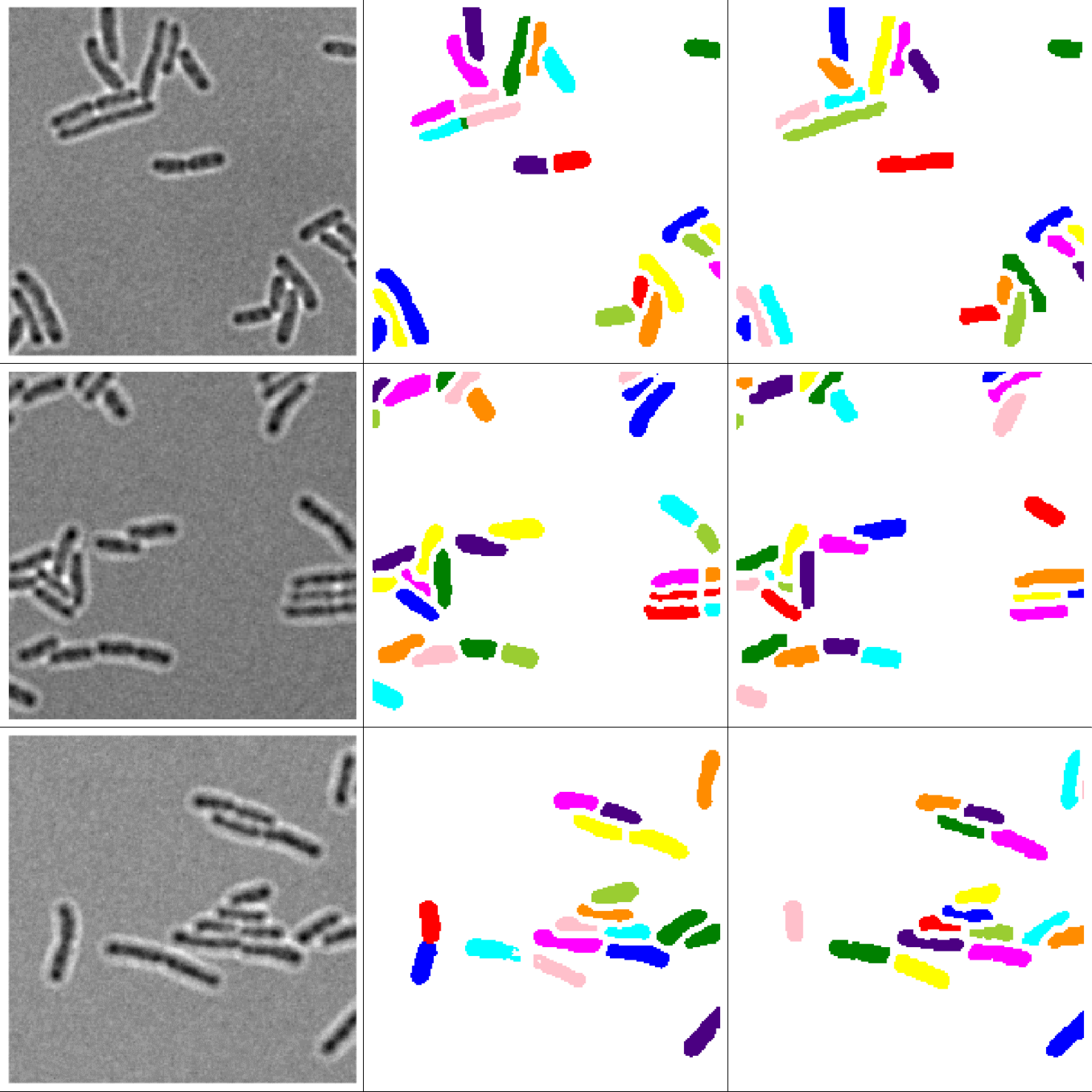}
\end{center}
\caption{ E.coli bacteria recorder under differential interference contrast microscopy. From left to right: raw image, result of our segmentation, ground truth labels. }
\label{fig:EColi}
\end{figure}

The E.coli dataset (Figure~\ref{fig:EColi}) is interesting because the number of organisms is large and they are crowded. The dataset contains 37 $1024\times1024$ bright-field images. The ground truth is derived using the watershed algorithm \cite{bieniek2000efficient} from weak annotations in which every organism is annotated by a line segment. The train-test splitting is done randomly selecting 30 as train images, and 7 as the test set. The validation is done on the train subset, so we have not used test set for the selection of meta-parameters.

At test time, images were processed by non-overlapping crops of size $256\times256$. The SBD score was calculated for each crop independently and then averaged. The performance of our method is better compared to other methods prevously evaluated on this dataset (Table \ref{tab:ecoli}). Unfortunately, we were not able to obtain reasonable results from the method~~\cite{de2017semantic}, probably due to a drastic change in the organism number between different crops. 
 
\begin{table}
    \begin{center}
        \begin{tabularx}{0.45\textwidth}{|X|c c|}
        \hline
        \textit{Method} &$\mid DiC\mid$&$SBD(\%)$\\
        \hline
        U-Net baseline \cite{ronneberger2015u} & - & 59.3\\
        Deep coloring \cite{kulikov2018instance}&2.2&61.9\\
        \hline
        \textbf{Ours}&\textbf{0.88}& \textbf{81.2}\\
        \hline
        \end{tabularx}
    \end{center}
    \caption{Results on the E.coli dataset. We follow the protocol from \cite{scharr2014annotated} to compare the results with those reported in \cite{ronneberger2015u} and \cite{kulikov2018instance}.}
    \label{tab:ecoli}
\end{table}

\subsection{HeLa dataset}

\begin{figure}
\begin{center}
\includegraphics[width=0.45\textwidth,keepaspectratio]{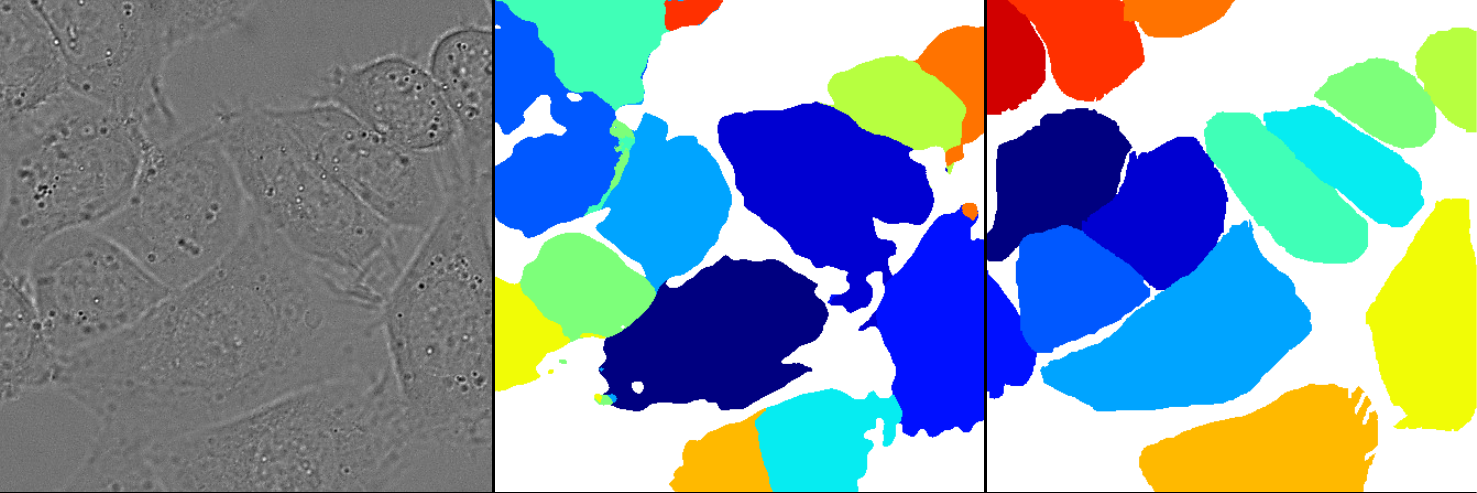}
\end{center}
\caption{ HeLa cells on glass recorded with differential interference contrast microscopy. From left to right: raw image, result of our segmentation, ground truth labels.  }
\label{fig:Hela}
\end{figure}

The HeLa cancer cells dataset\footnote{Courtesy Dr. Gert van Cappellen Erasmus Medical Center of Rotterdam.} (Figure \ref{fig:Hela}) is quite different from the other three datasets. Cells take a large part of each image and, being cancerous, are more irregular and form intricate patterns. In contrast to small and crowded pictures with E.coli, the number of cells is moderate, but they have a large area and more diverse sizes. The dataset contains 18 partially annotated single channel training images. Following best practices, we split the dataset into train and test parts (nine images each). The goal of this experiment is thus to show that our method can generalize well given very few training examples. 
 
We trained the network for 8000 epochs. No information about the background was used during training. On this dataset, we achieve \textbf{$78\%$} SBD without foreground mask, and $86\%$ SBD with foreground mask (imposed on top of the results), which slightly outperforms the semantic segmentation baseline of $77.5\%$ IOU reported in \cite{ronneberger2015u}. Our implementation of the baseline method~~\cite{de2017semantic} did not show any reasonable results with current configuration.   

\subsection{C.elegans dataset}

\begin{figure*}
\begin{center}
\includegraphics[width=\textwidth,keepaspectratio]{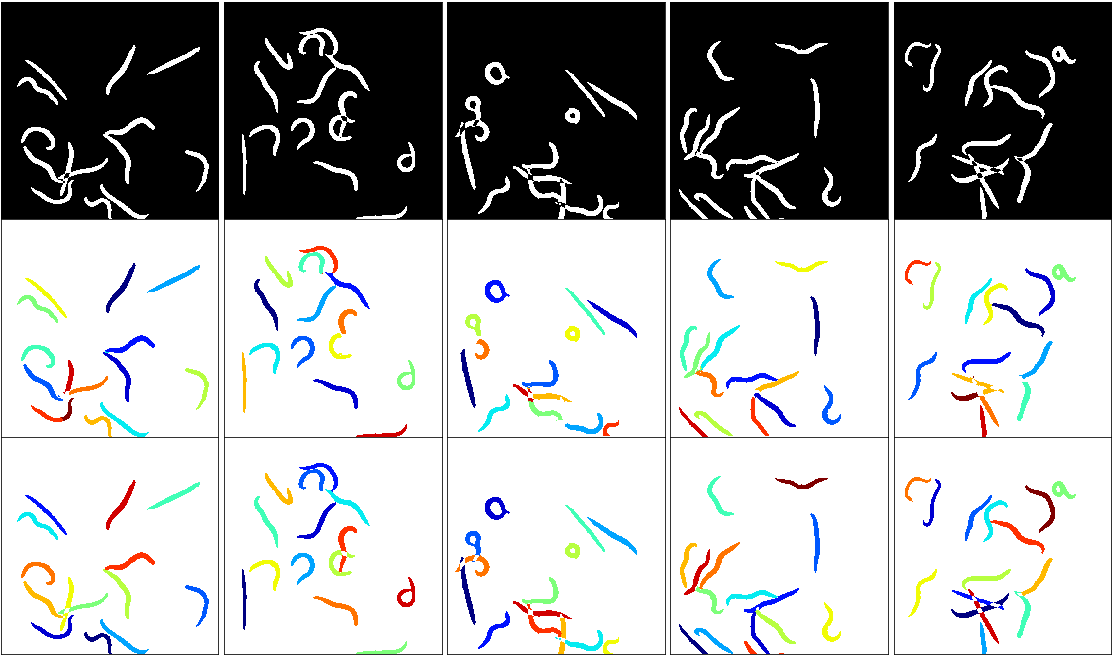}
\end{center}
\caption{ Binary images of C.elegans roundworm. Simple images crops (top row), second row - our results, third row - ground truth lables (color display needed). }
\label{fig:elegans}
\end{figure*}

Finally, we look at the C.Elegans dataset (Figure \ref{fig:elegans}), which is available from the Broad Bioimage Benchmark Collection \cite{ljosa2012annotated}. This sequence contains 97 two-channel images, $696 \times 520$ pixels each, of roundworm C.elegans. Each image contains approximately 30 organisms, some of them in complex overlapping patterns. In order to compare our results with those reported in  \cite{novotny2018semi} we followed their protocol: the entire dataset was split into  (two) equal parts: 50 training set and 47 test images. Here, we use the binary segmentation masks (following \cite{novotny2018semi,wahlby2010resolving,yurchenko2017parsing}). The network was trained for 1000 epochs. 

\begin{table*}
    \begin{center}
        \begin{tabularx}{0.9\textwidth}{|X|c c c c c|}
        \hline
        \textit{Method} & $AP$ & $AP_{0.5}$ & $AP_{0.75}$ & $AP_S$ & $AP_M$\\
        \hline
        Semi-convolutional operators  \cite{novotny2018semi} & 0.569 &0.885 &0.661 &0.511 &0.671\\
        Mask RCNN \cite{he2017mask} &0.559 &0.865 &0.641 &0.502 &0.650\\
        \hline
        Discriminative loss \cite{de2017semantic} (our implementation without SinConv)  & 0.343 &0.624 &0.380 &0.441 &0.563\\
        Discriminative loss \cite{de2017semantic} (our implementation with SinConv)  & 0.478 &0.771 &0.560 &0.551 &0.677\\
        \hline
        \textbf{Ours} & \textbf{0.724} & \textbf{0.900} & \textbf{0.723} & \textbf{0.775} & \textbf{0.875}\\
        \hline
        \end{tabularx}
    \end{center}
     \caption{Results on the C.elegans dataset. The results were obtained using the COCO standard metric \cite{lin2014microsoft}. Again, the proposed method outperforms previous approaches.} 
    \label{tab:worms}
\end{table*}

We evaluate the instance segmentation using average precision (AP) metric computed with the standard COCO evaluation protocol \cite{lin2014microsoft}. Table~ \ref{tab:worms} suggests that our method outperforms the previous works, including the well-known Mask-RCNN method~\cite{he2017mask} (as reported by  \cite{novotny2018semi}). 

\subsection{Ablation study}

We compare our method with the ablation that uses CoordConv in place of SinConv, as well as the ablations that use randomly initialized set of guide functions without optimization (three variants). Our unablated method is superior in all cases and across both datasets. In the case of Table \ref{tab:cvppp_coord2}, the advantage of our method over the baseline approach in  \cite{de2017semantic} is not significant. However, in other cases (Table \ref{tab:worms_coord2}) it is considerable. It is not surprising that the ablation of our method without SinConv struggles, as the desired function is very hard to learn using fully-convolutional operations. 
%It should not however, be seen as a disadvantage of our full pipeline.

\begin{table}[ht]
    \begin{center}
        \begin{tabularx}{0.49\textwidth}{|X|c c|}
        \hline
        \textit{Method} &$\mid DiC\mid$&$SBD(\%)$\\
        \hline
        SinConv&3.& 89.9\\
        \hline
        No guide &5.&78.3\\
        CoordConv & 3. & 87.6 \\
        \hline
        Random & 11. &70.8\\
        Low & 5. &86.2\\
        High& 16.9 &5.4\\
        \hline
        \end{tabularx}
    \end{center}
     \caption{The ablation study on the CVPPP dataset. We evaluate the variant without SinConv (standard convolutions -- row 2) and with CoordConv replacing SinConv (row 3). The last three rows are the variants with guide functions sampled from the uniform distribution. Random - indicates the whole range of frequencies, Low and High - sampled from low and high frequencies, respectively. The unablated variant of our method (top row) performs best.}
    \label{tab:cvppp_coord2}
\end{table}

\begin{table}[ht]
    \begin{center}
        \begin{tabularx}{0.49\textwidth}{|X|c c c c c|}
        \hline
        \textit{Method} & $AP$ & $AP_{0.5}$ & $AP_{0.75}$ & $AP_S$ & $AP_M$\\
        \hline
        SinConv & 0.724 & 0.900 & 0.723 & 0.775 & 0.875\\
        \hline
        No guide &	0.028&	0.063&	0.023&	0.105&	0.081\\
        CoordConv & 0.436 &	0.631&	0.414&	0.485&	0.685\\
        \hline
        Random & 0.078 &	0.115 &	0.080&	0.143&	0.179\\
        Low &	0.273&	0.498&	0.240&	0.350&	0.503\\
        High &	0.001&	0.002&	0.002&	0.010&	0.01\\
        \hline
        \end{tabularx}
    \end{center}
     \caption{The ablation study of our framework on the C.elegans dataset. Same ablations as in the previous table are considered. The advantage of our full method over ablations is greater for this more challenging task. }
    \label{tab:worms_coord2}
\end{table}

\begin{table*}[ht]
    \begin{center}
        \begin{tabularx}{0.9\textwidth}{|X|c c c c c c|}
        \hline
        \textit{Method} & $AP$ & $AP_{0.5}$ & $AP_{0.75}$ & $AP_S$ & $AP_M$& $AP_L$\\
        \hline
        Mask RCNN \cite{he2017mask} &0.319 &0.619 &0.309 &0.120 &0.365& 0.520\\
        \textbf{Ours}&0.223 &0.476 &0.194 &0.029 &0.238& 0.484\\
        \hline
        \end{tabularx}
    \end{center}
     \caption{Results on the COCO validation set (person class only). For this dataset with large scale variability, the results of our method are worse, especially for small objects.} 
    \label{tab:coco}
\end{table*}

\subsection{Method limitations}

Despite state-of-the-art results on biological datasets, the proposed method has several limitations that need to be resolved before applying it to more complex datasets with severe variation in object scale such as COCO \cite{lin2014microsoft}, PASCAL VOC \cite{pascal-voc-2012}, Cityscapes \cite{cordts2016cityscapes} (where our initial attempts to apply the method lead to mediocre, i.e.,\ mid-table results). To the best of our understanding, the sub-par performance of the method is caused by the inability to handle very diverse scales gracefully.

% We are currently investigating multi-scale schemes as well as other families of guide functions, which may potentially improve the results.

To be specific, we have compared our method with Mask RCNN \cite{he2017mask} implementation in PyTorch 1.2 \cite{pytorch} on MS COCO dataset \cite{lin2014microsoft} using our setup. We have used the validation part of the dataset; all images were transformed to the $448 \times 448$ resolution by down-sampling and zero-padding along the smaller dimension. The evaluation was performed on a single class: person (class=1). In this setting, the results of our methods are worse that those obtained using Mask RCNN (Table \ref{tab:coco}).  
\section{Conclusions}
% We have developed a new method for instance segmentation of biological objects using \textit{harmonic instance embedding}. 

% In a nutshell, for each instance pixel embedding is the expectation over a range of predefined harmonic guided functions. So a deep convolutional network can be trained end-to-end by minimizing the $L_1$ distance between its prediction and the ground truth embedding, implicitly reducing the variance inside the object embeddings.  To ease the learning process, we have introduced a SinConv layer, that propagates the underline harmonic function space in the decoder pathway of the neural network. To retrieve instances from embeddings, we use the Mean-Shift algorithm. Our method does not require any metric learning and can be suitable for a wild range of datatypes. Moreover, our approach does not require complex object interaction during the training and can learn successfully with only one object on the image.

% The approach was shown to well four bio-medical datasets (CVPPP, C.Elegans, E.Coli, HeLa), surpassing the current state-of-the-art. 

We have presented a new instance segmentation approach that exploits the peculiarities of biological images. The approach relies on a new type of embedding based on sine waves with parameters pre-adjusted to achieve separation of ground truth instances in the train dataset. We have shown that such precomputation of good embeddings greatly simplifies the learning stage whenever the same guide patterns are inputted in some of the convolutional layers of the embedding network. The ease of training is evidenced by the superior performance of our method compared to the original embedding-based approach proposed in \cite{de2017semantic}. 

We have shown that our method can handle rather diverse biological image data, while using the same relatively small architecture and the same set of meta-parameters. Such versatility is valuable for practical deployment  of the method to domain specialists.

\paragraph{Acknowledgements.}
This work was supported by the Skoltech NGP Program (MIT-Skoltech 1911/R). 
Computations were performed on Skoltech HPC cluster Zhores.

%\FloatBarrier
\ifnum\value{page}>8 \errmessage{Number of pages exceeded!!!!}\fi

{\small
\bibliographystyle{ieee}
\bibliography{refs}
}
\end{document}